\documentclass{article}

\usepackage[dblblindworkshop, nonatbib, final]{neurips_2025}

\usepackage[style=ieee,citestyle=numeric-comp,sorting=none]{biblatex}
\bibliography{vlm-tg-context}
\workshoptitle{First Workshop on Interpreting Cognition in Deep Learning Models (CogInterp)}

\usepackage[T1]{fontenc}    
\usepackage{hyperref}       
\usepackage{url}            
\usepackage{booktabs}       
\usepackage{amsfonts}       
\usepackage{nicefrac}       
\usepackage{microtype}      
\usepackage[table,xcdraw]{xcolor}         
\usepackage{graphicx}
\usepackage{caption}
\usepackage{subcaption}
\usepackage{listings}

\title{Context informs pragmatic interpretation in vision--language models}

\author{%
  Alvin W.M. Tan*,\enspace Ben Prystawski*,\enspace Veronica Boyce,\enspace Michael C. Frank \\
  Department of Psychology\\
  Stanford University\\
  Stanford, CA 94305 \\
  \texttt{tanawm@stanford.edu} \\
}

\begin{document}

\maketitle

\begin{abstract}
Iterated reference games---in which players repeatedly pick out novel referents using language---present a test case for agents' ability to perform context-sensitive pragmatic reasoning in multi-turn linguistic environments.
We tested humans and vision--language models on trials from iterated reference games, varying the given context in terms of amount, order, and relevance. 
Without relevant context, models were above chance but substantially worse than humans.
However, with relevant context, model performance increased dramatically over trials.
Few-shot reference games with abstract referents remain a difficult task for machine learning models.
\end{abstract}

\section{Introduction}

Recent advances in machine learning have produced multi-turn conversational agents, which are a public face of artificial intelligence \cite{openaiIntroducingChatGPT2024}.
The utility of conversational agents depends on capacities including natural language understanding, world knowledge, and instruction following \cite{guanEvaluatingLLMbasedAgents2025}.
Furthermore, success in multi-turn conversation requires the underlying language model to respond to the user's message appropriately given the preceding context, which in turn requires the language model to retain 
relevant contextual information and use it to interpret new messages. 

Such multi-turn interactions are a core feature of human communication: shared conversational history supports a shared system of semantic meaning \cite{clarkUsingLanguage1996, geurtsCommonGroundPragmatics2024}. 
\textit{Iterated reference games} are a common experimental paradigm for studying this type of communication. 
In these paradigms, a describer produces a description of a referent such that a matcher can correctly select the referent from a set of options. 
Games repeat over multiple rounds, typically resulting in the emergence of conventionalised referring expressions over repeated descriptions of the same referent \cite{clarkReferringCollaborativeProcess1986}.
These experiments demonstrate that humans dynamically adapt to their conversational partners, creating ad hoc context- and/or partner-specific meanings and conventions \cite{clarkReferringCollaborativeProcess1986, hawkinsPartnersPopulationsHierarchical2021, boyceInteractionStructureConstrains2024}.

Conversations---whether human--human or human--AI---require their participants to use contextual information to understand the meanings of utterances.
A system with the ability to do this would show two behavioural signatures: (1) the ability to interpret linguistic meaning pragmatically and (2) sensitivity to contextual information \cite{hawkinsVisualResemblanceInteraction2023, boyceIdiosyncraticNotOpaque2025, junkerSceneGramConceptualizingDescribing2025, gulCoGenLearningFeedback2024, huaTalkLessInteract2024, chenRetrospectiveLearningInteractions2025, sravanthiPUBPragmaticsUnderstanding2024}. 
In this work, we employ iterated reference games as a minimal test case to investigate whether state-of-the-art open vision--language models (VLMs) reason pragmatically about referring expressions given varying types of prior context, and how their sensitivity to context compares to that of humans.
Iterated reference games have previously been used to study AI systems for both generation and comprehension of descriptions, but this domain continues to prove challenging for AI systems, especially for generation with abstract images \cite{jiAbstractVisualReasoning2022, gulCoGenLearningFeedback2024, wangLVLMsAreBad2025}. 

\begin{figure}[th]
\centering
\includegraphics[width=\textwidth]{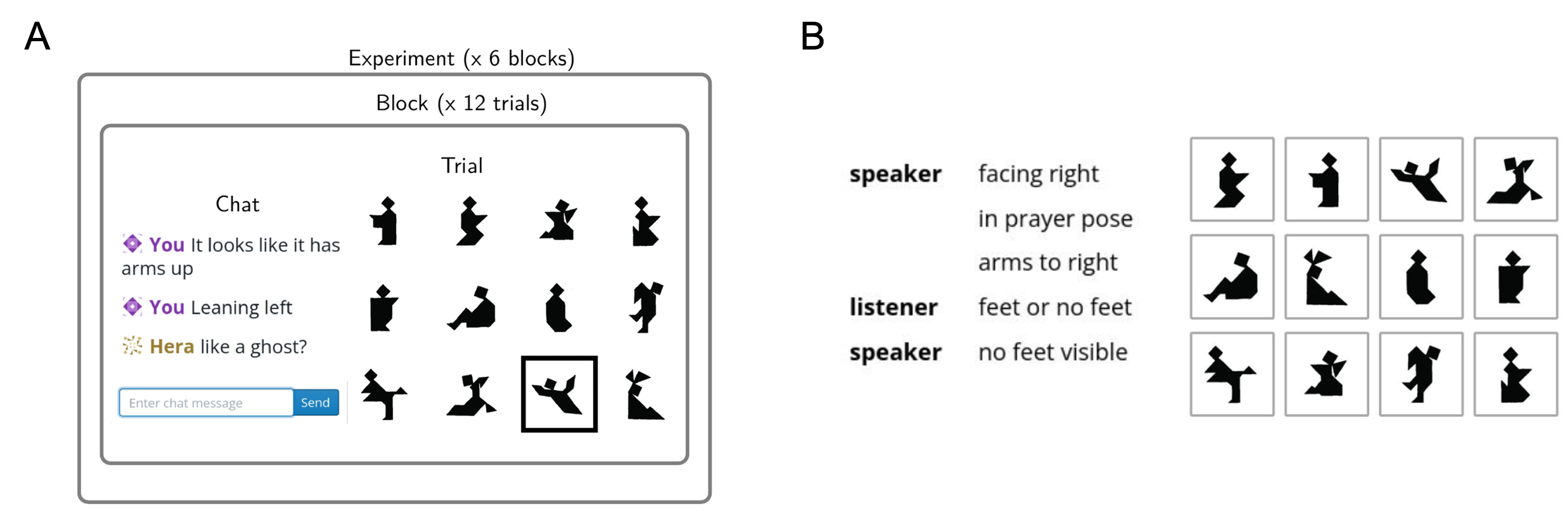}
\caption{A. Overview of the experimental structure for the original interactive games in Boyce et al.~\cite{boyceInteractionStructureConstrains2024}. B. User interface for each trial in the experiments with na\"ive participants. }
\label{fig:setup}
\end{figure}

\section{Methods}

All model data and code are available on Github at \url{https://github.com/benpry/vlm-tg-context/tree/coginterp}.

\subsection{Dataset}

We used an iterated reference game dataset from Boyce et al.~\cite{boyceInteractionStructureConstrains2024}. 
In each game, players saw a grid of 12 tangram images (Figure \ref{fig:setup}A).
On each trial, one player (the describer), saw one image highlighted. 
The describer described a target image to the other players via chatbox so that the matchers could each select the target image from the 12 options. 
Each game had 2--6 players, who played 6 rounds of 12 trials each (one trial for each image). 

Here we use ten of these games, which were subsequently used by Boyce et al.~\cite{boyceIdiosyncraticNotOpaque2025} to test for context sensitivity in naïve human matchers. 
Na\"ive matchers read transcripts of the conversations between the players in the original games and then made a selection of what image was being describer (Figure \ref{fig:setup}B). 
Na\"ive matchers received feedback on whether they were correct, but were not told the correct answer if their response was incorrect. 
Each participant saw all trials from one game, in an order that varied based on condition. 
Boyce et al.~\cite{boyceIdiosyncraticNotOpaque2025} collected participants in the yoked condition ($N$ = 99) and the shuffled condition ($N$ = 97). 
We collected additional human data on Prolific in the backward ($N$ = 89) and random ($N$ = 107) conditions (described in Section~\ref{sec:contexts}) using the same procedure.

\subsection{Experiment setup}

Our experiments with VLMs aimed to match the na\"ive human setup.
We evaluated the instruction-tuned versions of four leading open-weights VLMs of different sizes: Qwen 2.5 VL 32B \cite{bai2025qwen}, Gemma 3 27B \cite{team2025gemma}, Llama 3.2 11B \cite{llamateamLlama32Revolutionizing2024}, and Kimi VL A3B \cite{kimiteamKimiVLTechnicalReport2025}.
Models were given a general system prompt describing the task setup, followed by the set of 12 tangram image options labeled A through L presented as a single image. The exact prompt and image are included in Appendix~\ref{app:add_methods}.
Models were then given the context of all preceding trials formatted as a chat history, with the user supplying the text written by participants and the model serving as the assistant providing the target tangram, denoted A through L; note that models were given the correct target (unlike humans).
Finally, the description for the test trial was provided, and the model log probabilities for the tokens A through L were obtained.
We renormalised the log probabilities using a softmax, and treated the probability assigned to the correct target as the model's accuracy.

\subsection{Context conditions\label{sec:contexts}}
Do VLMs make use of context to the same extent, and in the same way, as humans? We considered eight conditions varying the amount and type of in-context trials seen by the model (Table~\ref{tab:conditions}).

\begin{table}[th]
    \small
    \centering
        \caption{Conditions of in-context trials seen by models. Columns indicate whether the in-context trials are from the same original game as the test trial, whether the in-context trials are all from the same original game, what the order of the in-context trials is, and whether the tangram in the test trial occurs in the in-context trials.}
        \label{tab:conditions}
    \begin{tabular}{lccccc}
\toprule
Condition    & $N_\textup{trials}$ & Original game             & Same game                 & Trial order                                             & Same tangram seen         \\ \midrule
Yoked        & 720                 & \cellcolor[HTML]{94F186}Y & \cellcolor[HTML]{94F186}Y & \cellcolor[HTML]{94F186}Original                        & \cellcolor[HTML]{94F186}Y \\
Shuffled     & 7056                & \cellcolor[HTML]{94F186}Y & \cellcolor[HTML]{94F186}Y & \cellcolor[HTML]{F18D86}Permuted                        & \cellcolor[HTML]{94F186}Y \\
Backward     & 720                 & \cellcolor[HTML]{94F186}Y & \cellcolor[HTML]{94F186}Y & \cellcolor[HTML]{F18D86}Reversed                        & \cellcolor[HTML]{94F186}Y \\
Ablated      & 720                 & \cellcolor[HTML]{94F186}Y & \cellcolor[HTML]{94F186}Y & \cellcolor[HTML]{94F186}Original & \cellcolor[HTML]{F18D86}N \\
Other-within & 6480                & \cellcolor[HTML]{F18D86}N & \cellcolor[HTML]{94F186}Y & \cellcolor[HTML]{94F186}Original & \cellcolor[HTML]{94F186}Y \\
Other-across & 6480                & \cellcolor[HTML]{F18D86}N & \cellcolor[HTML]{F18D86}N & \cellcolor[HTML]{94F186}Original & \cellcolor[HTML]{94F186}Y \\
Random       & 7704                & \cellcolor[HTML]{F18D86}N & \cellcolor[HTML]{F18D86}N & \cellcolor[HTML]{F18D86}Permuted & \cellcolor[HTML]{94F186}Y \\
No context   & 720                 & \cellcolor[HTML]{F18D86}N & \cellcolor[HTML]{F18D86}N & \cellcolor[HTML]{F18D86}None                            & \cellcolor[HTML]{F18D86}N \\ \bottomrule
    \end{tabular}
\end{table} 

The conditions varied in their fidelity to the context experienced by actual human participants in the original iterated reference games.
Four conditions contained trials drawn from a single original game. 
The \textit{yoked}, \textit{shuffled}, and \textit{backward} conditions presented all the trials in the original game, in the original order, a permuted order, or the reversed order respectively.
The \textit{ablated} condition presented the trials in the original order, but in-context trials containing the same target tangram as the test trial were removed.
Two other conditions involved in-context trials sampled from different original games than the test trial.
The \textit{other-within} condition drew all in-context trials from a single different original game than the test trial, whereas the \textit{other-across} condition drew in-context trials randomly from all other original games.
We also included a \textit{random} condition that fully shuffled across games and trial orders.
Finally, we had a baseline condition of \textit{no context}, in which no in-context trials were presented.

\section{Results}

\begin{figure}[th]
    \centering
    \includegraphics[width=\linewidth]{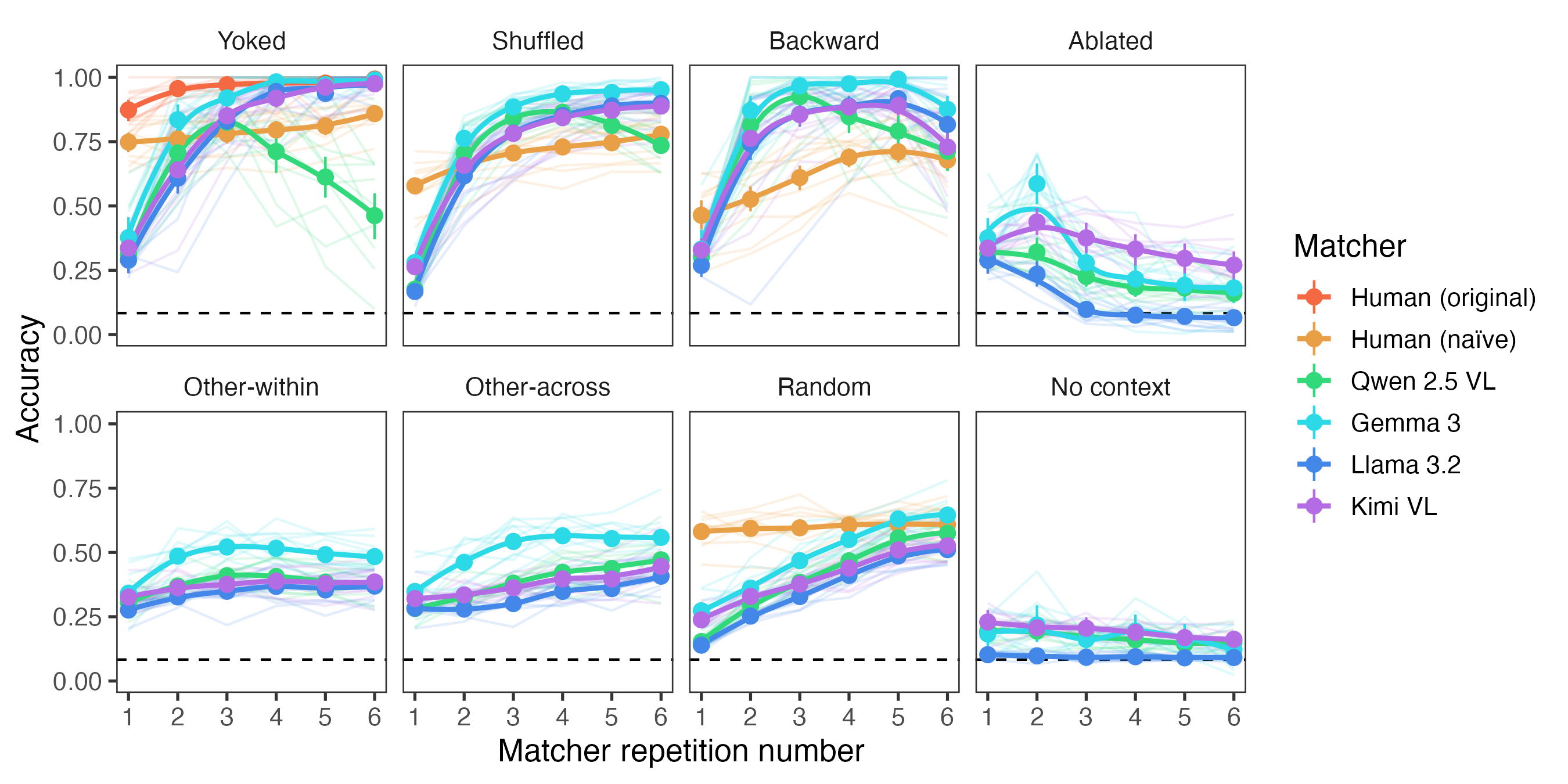}
    \caption{Matcher accuracy across all conditions and matcher types (both human and model), with best-fit LOESS curves, shown by repetition number as seen by the matcher, except for the no context condition where repetition number is from the original game. Error bars indicate bootstrapped 95\% confidence intervals. Dashed lines indicate the chance level (0.083). }
    \label{fig:matcher_plot}
\end{figure}

Our first question was whether models are able to perform ad hoc pragmatic reference resolution in this abstract, out-of-domain context. 
Results from all models and conditions, along with comparative human data from the original reference games \cite{boyceInteractionStructureConstrains2024} and from na\"ive humans \cite{boyceIdiosyncraticNotOpaque2025}, are shown in Figure~\ref{fig:matcher_plot}.
All models demonstrated above-chance accuracy on the task across all conditions, but without context, models were only slightly above chance. 
In the random condition, model performance improved as the number of in-context examples increased, suggesting some adaptation to the task. In contrast, na\"ive humans' performance was relatively high even in the first repetition and remained steady over time. 
In the absence of conversational history, models appear to have much worse ``intuition''  for interpreting pragmatic references to tangram images than humans. 

Our second question was how models would perform with relevant conversational history.
The yoked, shuffled, and backward conditions show how model performance was affected by the \textit{amount} and \textit{order} of contextual information. Performance in the shuffled condition was generally poorer than in the yoked condition, matching the trend of na\"ive humans.
Models were more accurate in the backward condition than in the shuffled condition, unlike humans, suggesting that working backwards from conventions to earlier and less conventionalised expressions may be easier for models than for humans. 
While humans outperformed models on the first trials seen, models were able to rapidly exploit the conversational history to achieve $\sim$0.8 accuracy.
These disparities suggest that VLMs may be worse at zero- or few-shot novel task performance than humans, but are able to learn relatively quickly with sufficient examples. 

Within-game context was critical to models. 
When we varied the \textit{relevance} of the context by drawing the conversational history from a different game or games (other-within and other-across), models had much lower accuracy of 0.3 -- 0.5. 
This lower performance compared to yoked, shuffled, and backwards suggests that model improvement with relevant context is not simply adaptation to the task. Rather, the boost depends on the context and the test trial coming from the same original game. The convention reached by one game is not necessarily predicted by the context from other games. 
Performance was also low in the ablated condition, implying that experience with the test trial tangram itself is crucial to understand a convention, which is not systematically inferrable from the other tangram conventions, even from the same game. 

\begin{figure}[th]
    \centering
    \includegraphics[width=\linewidth]{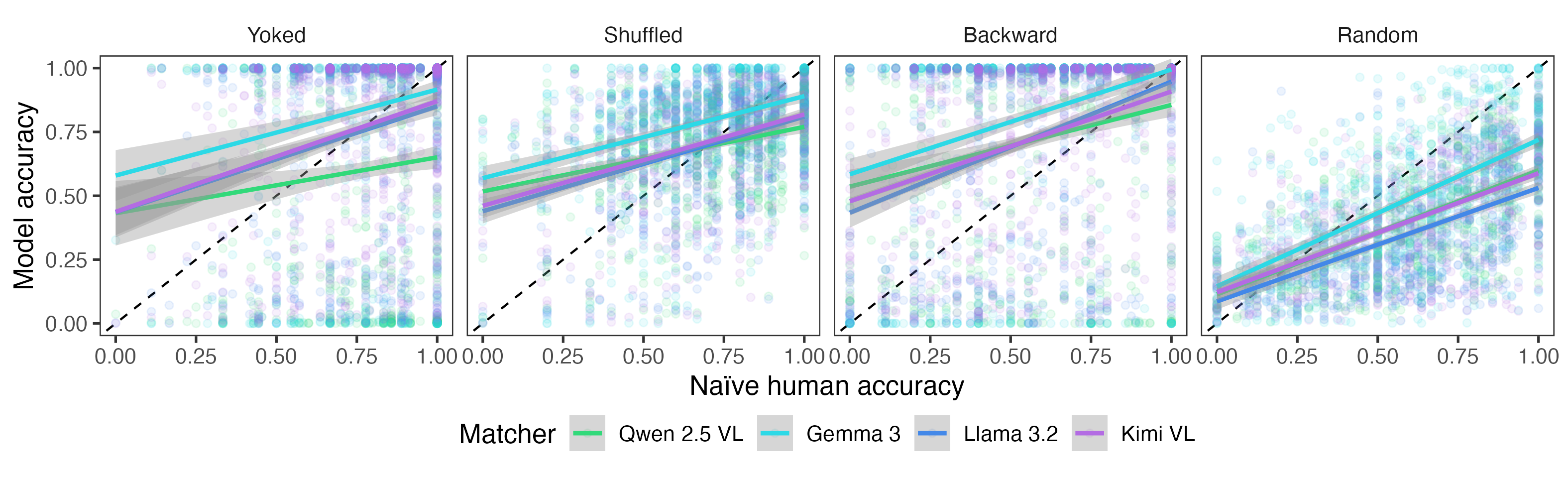}
    \caption{Comparison between naïve human accuracy and model accuracy, with best-fit linear regressions. Shaded regions indicate bootstrapped 95\% confidence intervals. Dashed lines indicate perfect calibration ($y = x$).}
    \label{fig:comparison}
\end{figure}

To determine whether models generally show the same pattern of difficulties as humans, we also compared model and na\"ive human performance at the trial level  (Figure~\ref{fig:comparison}).  
Models were relatively poorly calibrated, with weak correlations between model and human performance (yoked model $r$ = .10 -- .27, human split-half $r$ = .42 [.32, .50]; backward model $r$ = .23 -- .40, human split-half $r$ = .48 [.40, .56]; see Appendix~\ref{app:corrs} for more details).
Models may not have the same factors affecting their performance as humans, resulting in a lack of trial-wise similarity. 

Across models,  Gemma 3 was the best performing model across most conditions. 
Interestingly, Qwen 2.5's performance diverged from the other VLMs in the yoked, shuffled, and backward conditions, performing much worse in later rounds; we speculate this finding is due to the longer context lengths in later rounds, which Qwen 2.5 may be worse at handling.

Additional analyses on item-level effects, condition-wise correlations, and errors are shown in Appendix~\ref{app:add_analyses}, and additional results from a no-image, text-only paradigm are shown in Appendix~\ref{app:add_results}.

\section{Discussion}

Multi-turn conversation is a key aspect of human communication and a challenge for AI agents.
We demonstrated that state-of-the-art open-weights VLMs were able to perform ad hoc pragmatic reasoning across multiple turns of a reference game. 
We found that models were sensitive to the amount, order, and relevance of contextual information, showing rapid improvement when in-context examples came from the same source game and included the target image. 
This performance reflects a capacity for flexible meaning construction in a range of different contexts.

Nonetheless, models displayed different patterns of performance than humans and showed weak trial-wise calibration to human accuracies. 
Our results align with previous work on in-context learning, where increasing the number of in-context examples improved model performance, especially on difficult tasks \cite{chenHowManyDemonstrations2023, agarwalManyShotInContextLearning2024, jiangManyShotInContextLearning2024}.
While humans do show practice effects, these effects are typically much smaller.
The differential sensitivity to quantity and type of context between humans and models could suggest different approaches to the task. 
For example, models can (theoretically) exactly retrieve previous trial descriptions, whereas humans have more limited working memories and may need to use compression or other strategies to recall appropriate description--referent mappings. 

While we tested humans and models on the same stimuli and some of the same stimulus orders, the models received more informative feedback than humans. Models received the correct answers for the in-context examples, while humans only learnt if their selections were correct; this disparity may have contributed to differences in performance (see Section~\ref{app:learning}). 
Additionally, we tested a limited set of abstract stimuli as a representative example; generalisability to other (e.g., naturalistic) stimuli sets is left to future work. 

Overall, reference games with abstract referents remain a difficult task for machine learning models, particularly in the few-shot setting \cite{gulCoGenLearningFeedback2024, chenRetrospectiveLearningInteractions2025}. 
Our results point to the important role of contextual information, but also highlight ways in which current VLMs are overly sensitive to context.
Further work is needed to more comprehensively characterise the role of prompt engineering, causal and attentional factors underlying context use, and humans' own sensitivity to context.
Furthermore, our study focused on interpretation, but generating appropriate descriptions is a yet harder task that is worth exploring.
These directions will help us to build language models that are flexible but robust, and that can adapt to a range of tasks and settings as humans do.


\printbibliography



\appendix

\section{Additional methodological details}\label{app:add_methods}

\subsection{Prompt}

The following text was used as the system prompt for all language model generations.

\begin{lstlisting}
You will be presented with a list of messages between people playing a reference game, where the describer has to get the matcher to choose a shape from a set of shapes. Your goal is to guess which of the shapes the describer is trying to get the matcher to choose. The shapes, with their labels, are shown in the image.
Please answer with just the letter corresponding to the image you think the describer is trying to get the matcher to choose.
\end{lstlisting}

\begin{figure}[th]
\centering
    \includegraphics[width=.3\linewidth]{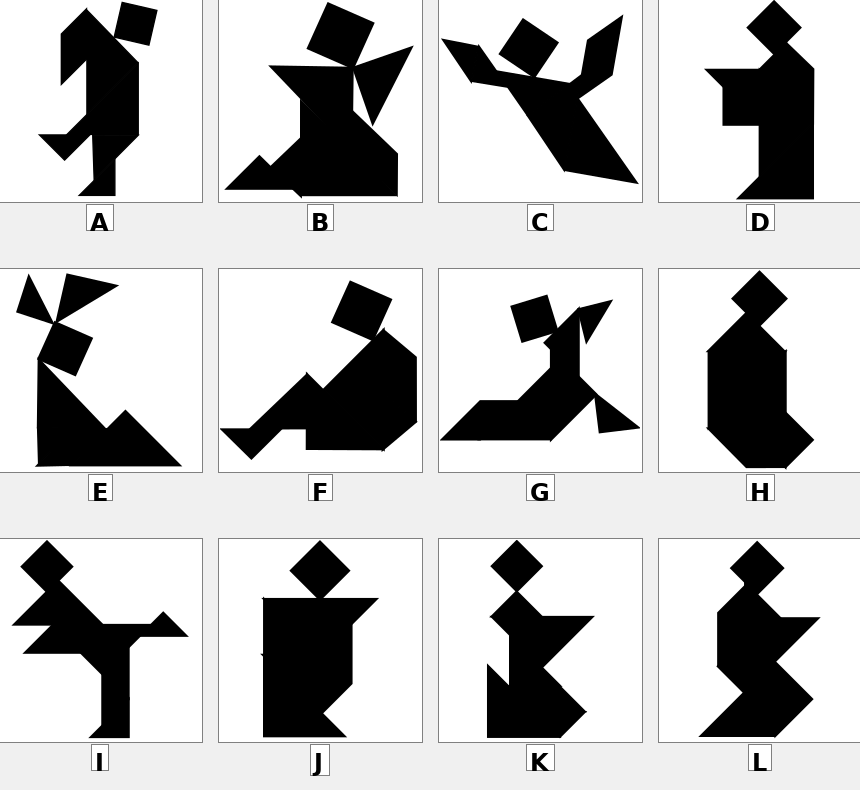}
    \caption{The image that was presented to the vision-language models, containing all 12 tangram shapes with their letter labels.}
\label{fig:tangrams}
\end{figure}

The image shown in Figure~\ref{fig:tangrams} was included with this prompt.

\section{Additional analyses\label{app:add_analyses}}
\subsection{Role of original trial number}

\begin{figure}[th]
    \centering
    \includegraphics[width=0.63\linewidth]{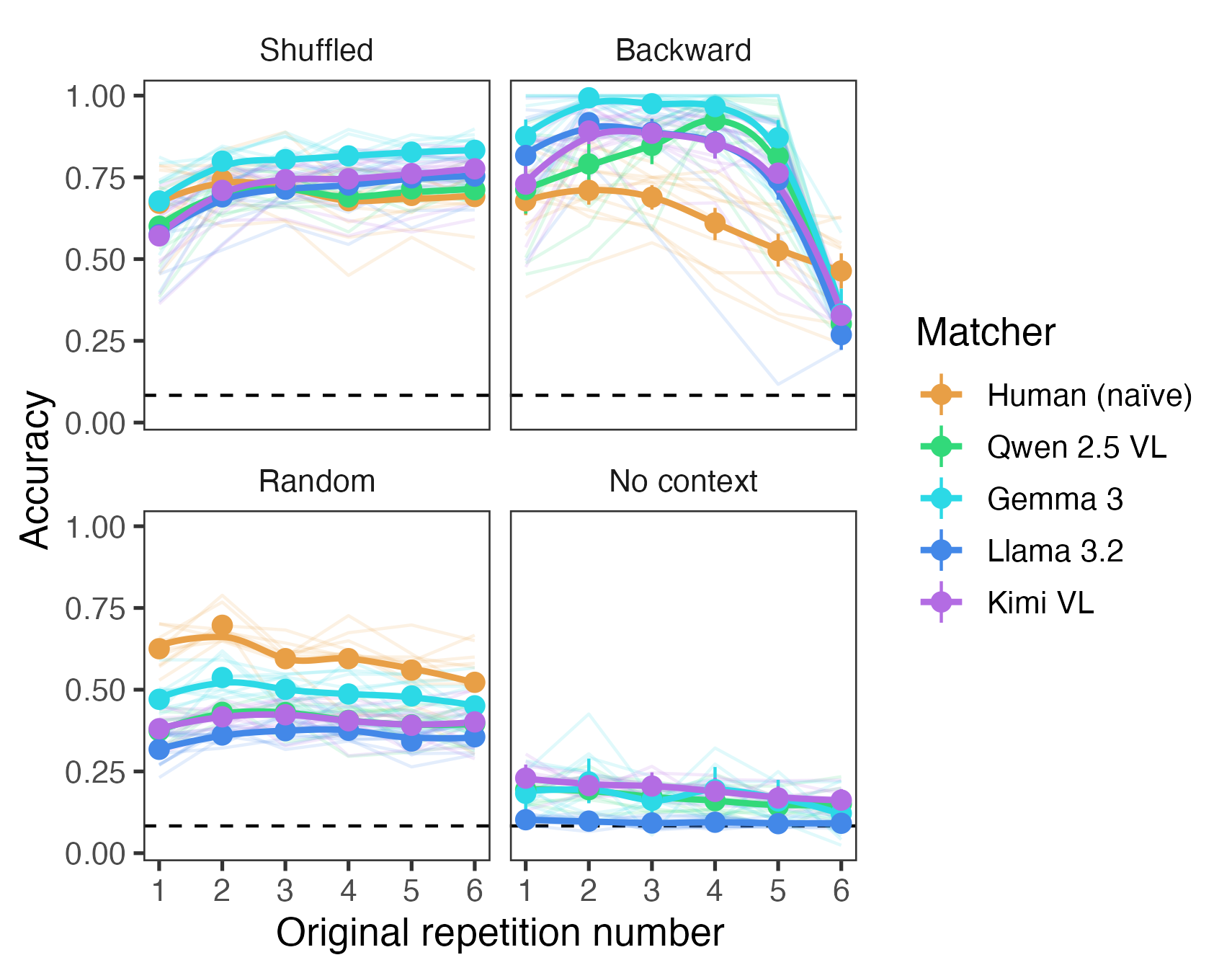}
    \caption{Matcher accuracy for the shuffled, backward, random, and no context conditions across all matcher types (both human and model), with best-fit LOESS curves, shown by repetition number from the original game. Error bars indicate bootstrapped 95\% confidence intervals. Dashed lines indicate the chance level (0.083). }
    \label{fig:original_plot}
\end{figure}

Four conditions displayed trials in a different order than the original games: shuffled, backward, random, and no context.
To understand the effect of the original trial number, we replotted matcher performance for these conditions by original repetition number in Figure~\ref{fig:original_plot}; note that the results for the no context condition are identical to those in Figure~\ref{fig:matcher_plot} as they were shown by original repetition number in that plot.
The general worsening performance over rounds across these conditions suggests that increasing conventionalisation in human referring expressions leads to more idiosyncratic expressions (as in \cite{boyceIdiosyncraticNotOpaque2025}) that are harder to resolve.
Interestingly, humans (and some models) showed highest performance in the second repetition (rather than the first).
This phenomenon suggests that there might be a trace of practice effects in human describers, whereby they exhibit increasing systematisation from the first to the second repetition, before exhibiting conventionalisation of expressions.

\subsection{Model--human comparisons\label{app:corrs}}

\begin{table}[th]
    \centering
    \caption{Human split-half correlations and model--human correlations for the yoked, shuffled, backward, and random conditions. Human split-half correlations show the mean and 95\% confidence interval over 1000 random splits.}
    \label{tab:corrs}
    \begin{tabular}{lccccc}
    \toprule
    & \multicolumn{5}{c}{$r$} \\ \cmidrule{2-6}
    Condition & Human split-half   & Qwen 2.5 VL & Gemma 3 & Llama 3.2 & Kimi VL \\ \midrule
    Yoked     & .42 {[}.32, .50{]} & .10         & .20     & .25       & .27     \\
    Shuffled  & --- & .28         & .34     & .38       & .37     \\
    Backward  & .48 {[}.40, .56{]} & .23         & .31     & .40       & .35     \\
    Random    & --- & .61         & .58     & .55       & .57     \\ \bottomrule
    \end{tabular}
\end{table}

Correlation values between models and na\"ive humans, as well as human split-half correlations, are shown in Table~\ref{tab:corrs}.
For the shuffled and random conditions, we averaged accuracies for each original trial prior to calculating correlations.
Human split-half correlations were computable only for yoked and backward conditions, in which multiple participants saw the same trials in the exact same order.
We also note that the human split-half correlations are likely underestimated as some splits did not have the whole set of original trials in each split, and thus had some dropped trials.
Overall, models were less similar to humans than humans were to other humans in the yoked and backward conditions; we hypothesise that the same would be true for the shuffled and random conditions if appropriate data were available.

\subsection{Item-wise variation}
\begin{figure}[th]
    \centering
    \includegraphics[width=\linewidth]{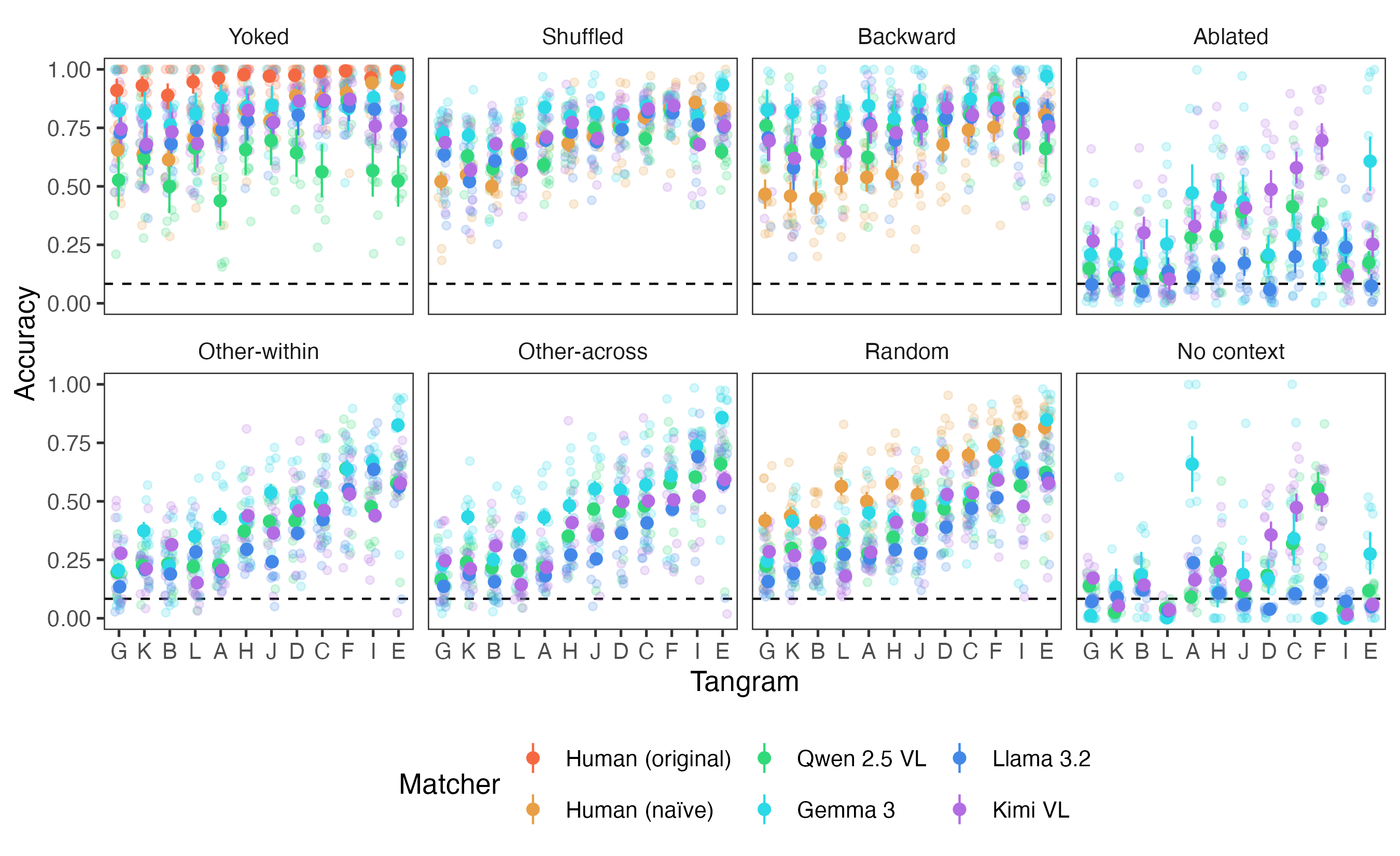}
    \caption{Item-wise variation in accuracy across matchers and conditions. Error bars indicate bootstrapped 95\% confidence intervals. Dashed lines indicate the chance level (0.083).}
    \label{fig:item_plot}
\end{figure}

Figure~\ref{fig:item_plot} shows the item-wise performance for all matchers and conditions. 
Figure~\ref{fig:tangrams} provides the grid of tangrams shown to the models, with correspondence between tangrams and letter labels. 

Both naïve humans and models display large item-wise variation, with the magnitude of this variation sometimes exceeding that due to repetition number, matcher, or condition, replicating the item-wise variation found by Boyce et al.~\cite{boyceIdiosyncraticNotOpaque2025}.
Item-wise variation appears to be largest for the other-within and other-across conditions, suggesting that some tangrams may be relatively more nameable or distinguishable than others, allowing for higher accuracy even without coherent context.
There appears to be some shared variance across models, such that the mean accuracy for each item is somewhat consistent across models for each condition, although each model also has particular idiosyncrasies and biases (e.g., Gemma 3 appears to be especially accurate for tangrams A and E compared to other models).

\subsection{Learning analyses\label{app:learning}}

\begin{figure}
    \centering
    \includegraphics[width=\linewidth]{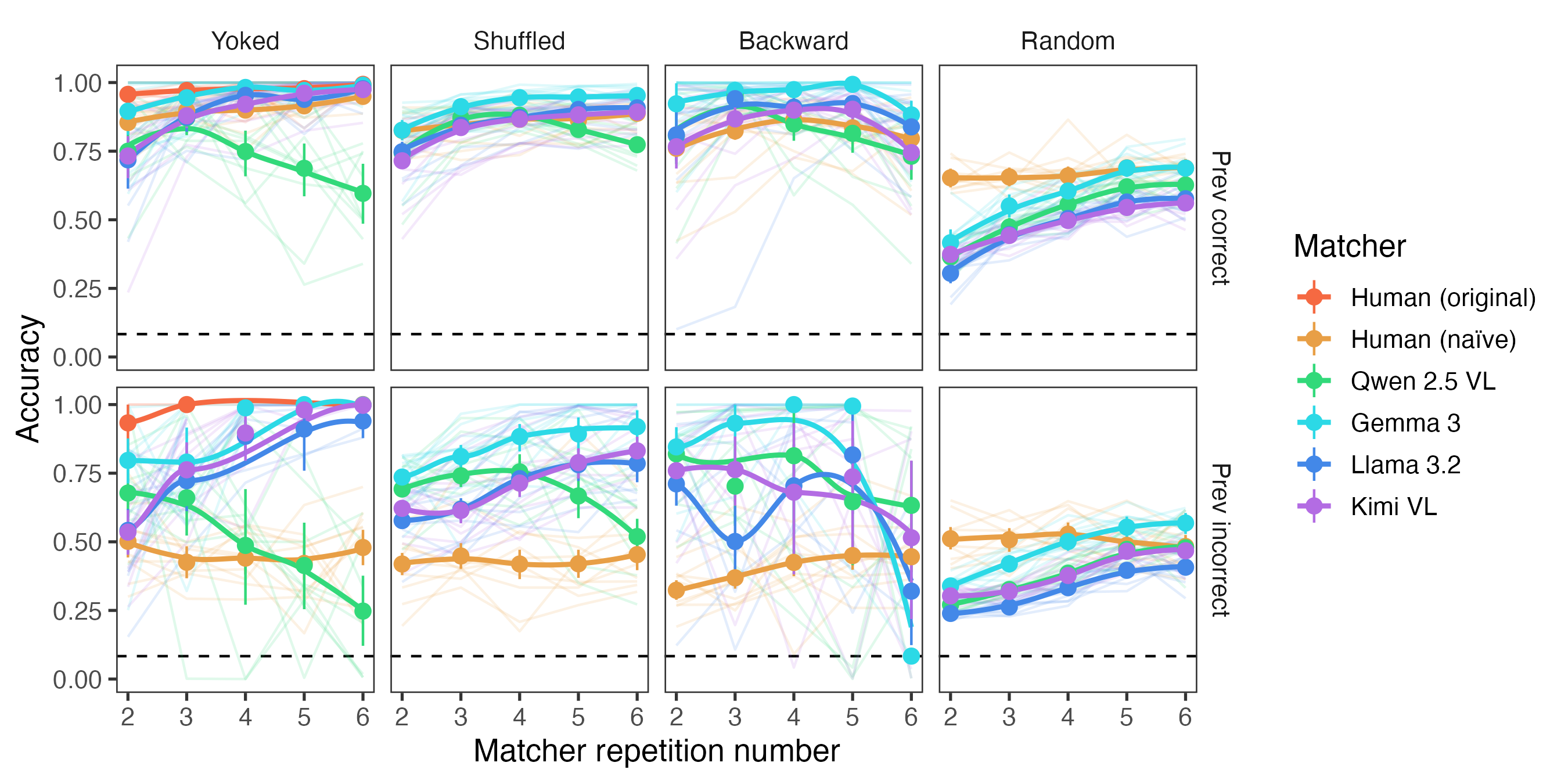}
    \caption{Trial accuracy as a function of previous-repetition accuracy for the same tangram across matchers for the yoked, shuffled, backward, and random conditions. Error bars indicate bootstrapped 95\% confidence intervals. Dashed lines indicate the chance level (0.083).}
    \label{fig:learning}
\end{figure}

To investigate the potential role of feedback, we visualised matcher accuracy from the second repetition onward as a function of whether the matcher selected correctly in the previous repetition of the same tangram, as shown in Figure~\ref{fig:learning}.
We quantised previous repetition accuracy by determining whether the argmax of the probabilities was the target tangram for models, and by determining whether any matcher selected correctly for the original-game humans; na\"ive human performance was already binarised.

The random condition replicated the overall accuracy curves in Figure~\ref{fig:matcher_plot}; since the prior context did not provide any relevant cues as for subsequent test trials, both models and humans performed similarly regardless of whether the previous trial was correct or not.

In the yoked, shuffled, and backward conditions, for both humans and models, when matchers answered a previous repetition correctly, there were also very likely to answer the next repetition correctly, with accuracy > 0.7 (except for Qwen 2.5 in the yoked condition).
However, the pattern of results was different when matchers answered a previous repetition incorrectly.
In this case, models were generally able to learn from previous mistakes in the yoked and shuffled condition (again, except for Qwen 2.5), whereas na\"ive humans were at about 0.5 accuracy---in other words, they were able to correct their previous mistakes only approximately half the time.

Some of this discrepancy may have been due to the fact that models received the correct target for all previous trials, whereas na\"ive humans only received feedback about whether they had selected correctly (i.e., they knew the correct tangram only when they had selected it correctly themselves).
Future work can more fully characterise the role of feedback by either allowing na\"ive humans to know the correct tangram if they answered incorrectly, or by presenting models with only selection correctness feedback.
We note also that the construct of ``learning'' is slightly disanalogous between humans and models, since humans experienced trials sequentially whereas models did not.
Evaluating models in a true multi-turn setting would help to more closely align the experiment setups.

\subsection{Correlation analyses}

\begin{figure}
    \centering
    \includegraphics[width=\linewidth]{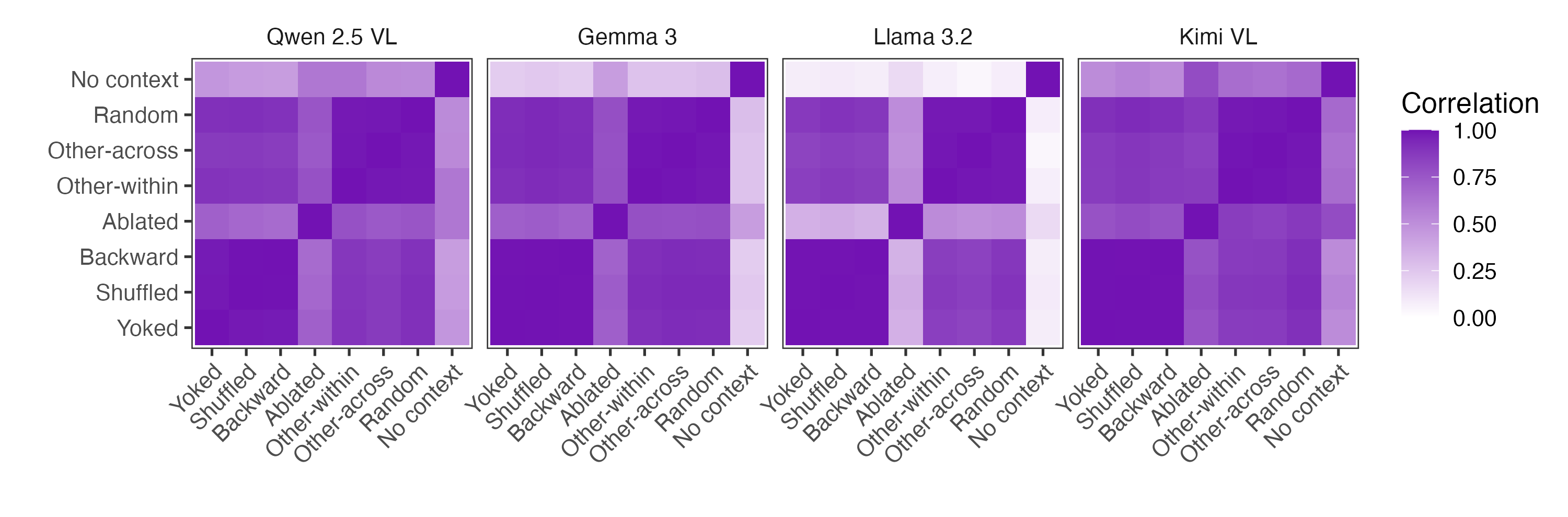}
    \caption{Correlation in confusion matrices across conditions for all models.}
    \label{fig:rsa}
\end{figure}

Drawing inspiration from representational similarity analysis \cite{kriegeskorteRepresentationalSimilarityAnalysis2008}, we conducted a correlation analysis over the confusion matrices for each model across conditions.
We constructed confusion matrices for each model--condition combination by taking the mean of the probability distributions across trials for each target tangram.
We then calculated the Pearson's correlation among the confusion matrices for all conditions.
The resultant correlograms are shown in Figure~\ref{fig:rsa}.
These correlograms demonstrate relatively high correlations across conditions for all models (all $r > .83$), except for the ablated condition ($r$ = .16 -- .87) and the no context condition ($r$ = .04 -- .80).
Furthermore, all models show a similar pattern of correlations, with the yoked, shuffled, and backward conditions clustering together, the other-within, other-across, and random conditions clustering together, and the ablated and no context conditions being the most distinct from the other conditions.
This pattern of results partly reflects the differences in overall performance across the different conditions, but also supports the idea that relevance is a crucial organising dimension of the content of contextual information.

\subsection{Error analyses}
\begin{figure}
    \centering
    \includegraphics[width=0.8\linewidth]{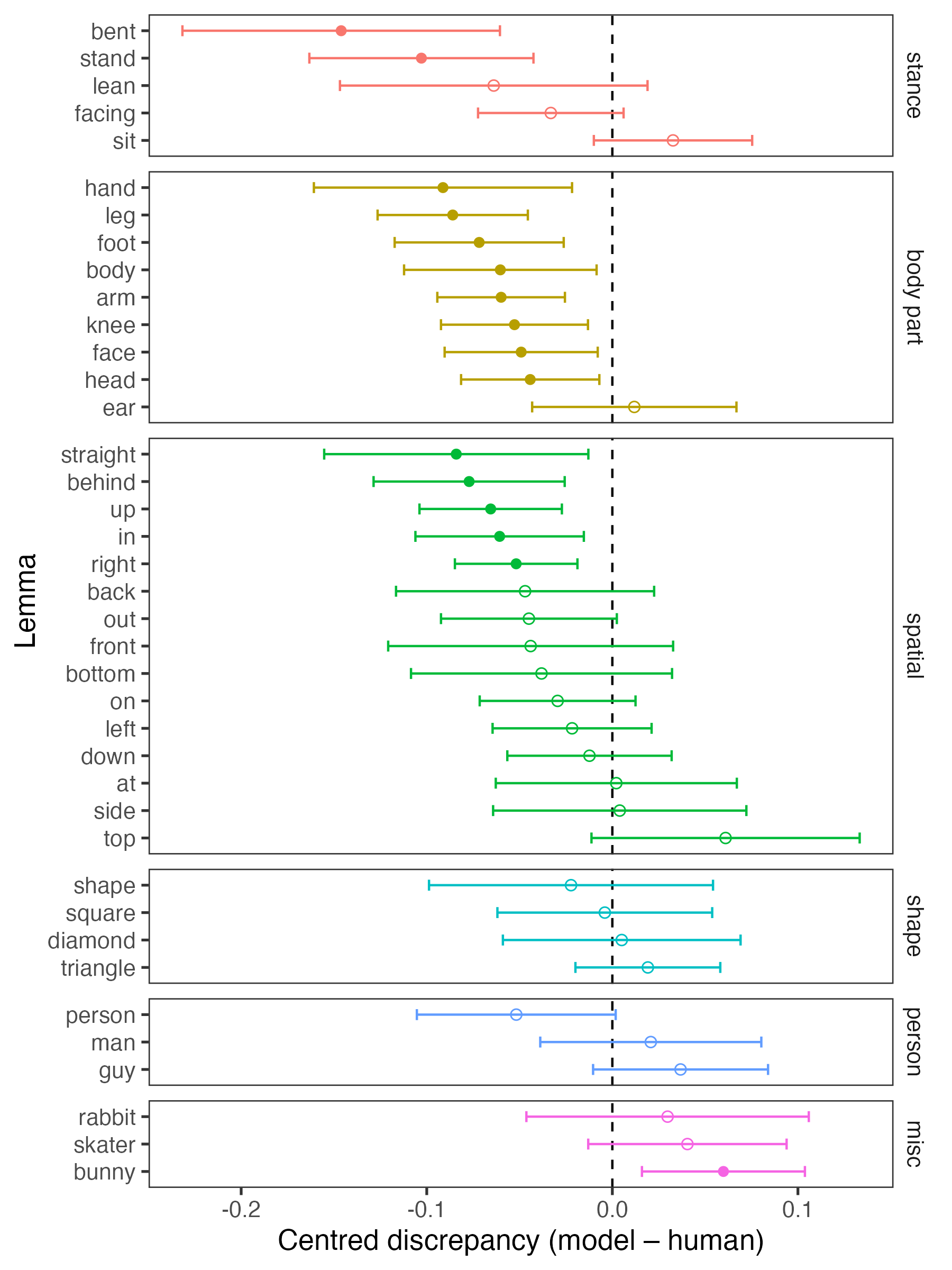}
    \caption{Centred discrepancy in model versus human accuracies in the random condition for selected lemmas, categorised by semantic categories. Error bars indicate 95\% confidence intervals. Dashed line indicates zero (i.e., discrepancy equal to the grand mean). Empty/filled points indicate whether the confidence intervals do/do not include zero respectively.}
    \label{fig:errors}
\end{figure}

We conducted an error analysis to investigate possible factors driving differences between model and human performance.
We used model and na\"ive human responses in the random condition, in which there was limited relevant context, although we note that our method could be applied to other conditions as well.
We estimated the discrepancy between model and human performance by calculating the difference between model and human accuracy for each unique trial, averaging across all models and runs.
Then, we lemmatised the message texts in each trial, and estimated the means and 95\% confidence intervals of the model--human discrepancy for each unique lemma across all trials in which that lemma occurred.
We recentred these discrepancy scores by subtracting the grand mean in discrepancy across all trials (-0.186), and determined the significance of the centred discrepancy (CD) estimates by calculating whether the 95\% confidence intervals contained zero. 
These CD values reflected whether models were better or worse than average, in comparison to humans.
We also filtered down to lemmas that occurred in at least 4 games and in at least 20 trials (i.e., their discrepancies were not due to a small number of idiosyncratic games).
The CD values for selected lemmas are shown in Figure~\ref{fig:errors}.

Manual inspection of the resulting estimates suggested that there were very few lemmas for which centred discrepancy was positive; the only such lemma that occurred in at least 4 games was ``bunny''. 
Conversely, there were 39 lemmas for which centred discrepancy was negative and which occurred in at least 4 games.
Two notable groups of such lemmas were body parts (e.g., ``hand'', ``leg'', ``foot'', ``arm''), and some prepositions and direction words (e.g., ``behind'', ``up'', ``right''). 
Together, these discrepancies suggest that models were worse than humans at interpreting non-literal tangram parts and their relations.
Intriguingly, models were not worse than average for several general person-related words (e.g., ``person'', ``man'', ``guy''), shape words (e.g., ``shape'', ``triangle'', ``square'', ``diamond''), and some other spatial terms (e.g., ``left'', ``down'', ``side'', ``top''). 
This contrast suggests that models were average at interpreting more literal tangram parts and their relations, as well as more general or holistic humanoid descriptions.

\section{Additional results\label{app:add_results}}
\subsection{Image--text vs. text-only performance}

\begin{figure}
    \centering
    \includegraphics[width=\linewidth]{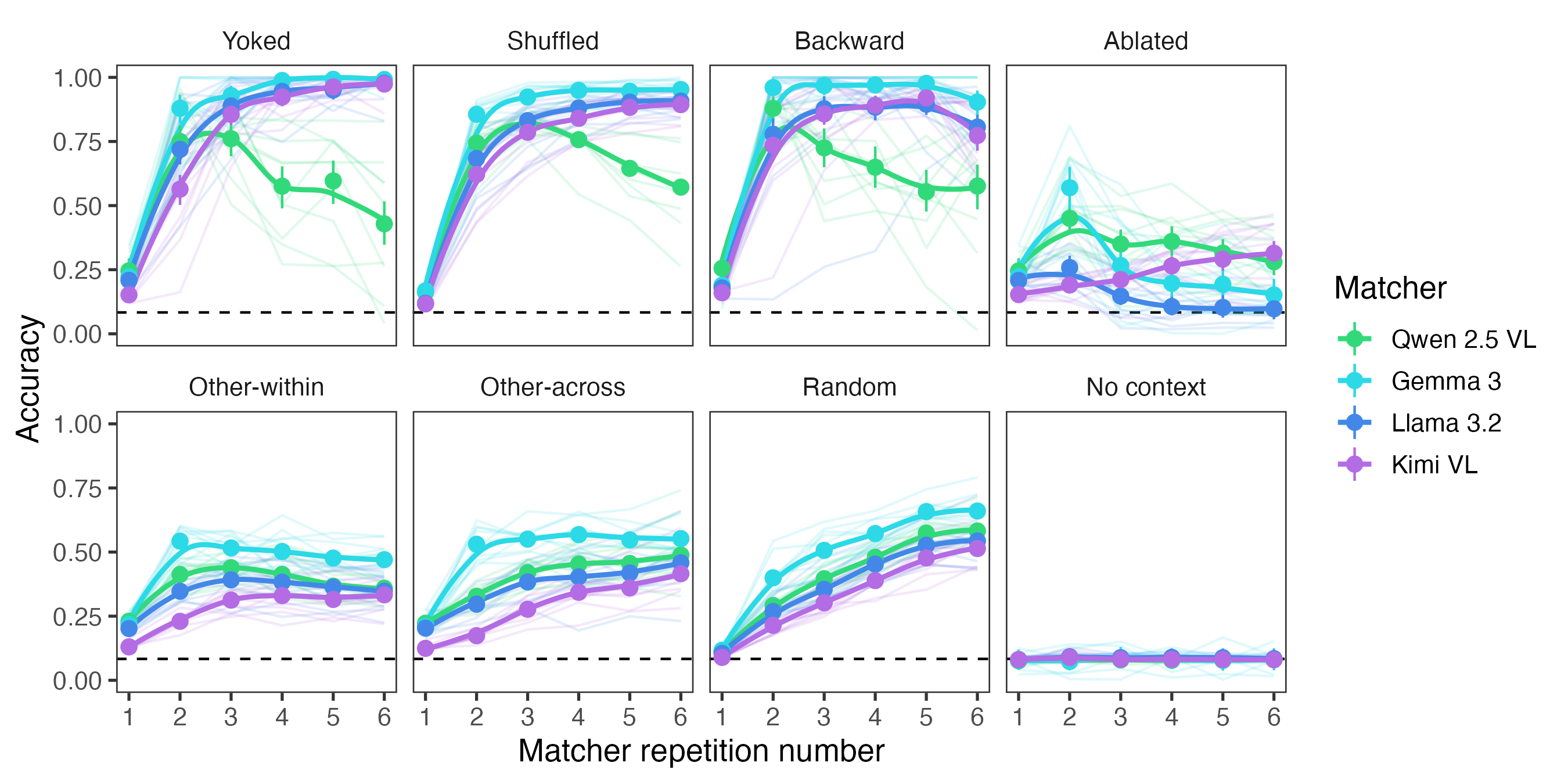}
    \caption{Matcher accuracy across all conditions and models using text input only, with best-fit LOESS curves, shown by repetition number as seen by the matcher, except for the no context condition where repetition number is from the original game. Error bars indicate bootstrapped 95\% confidence intervals. Dashed lines indicate the chance level (0.083).}
    \label{fig:noimg}
\end{figure}

\begin{figure}
    \centering
    \includegraphics[width=\linewidth]{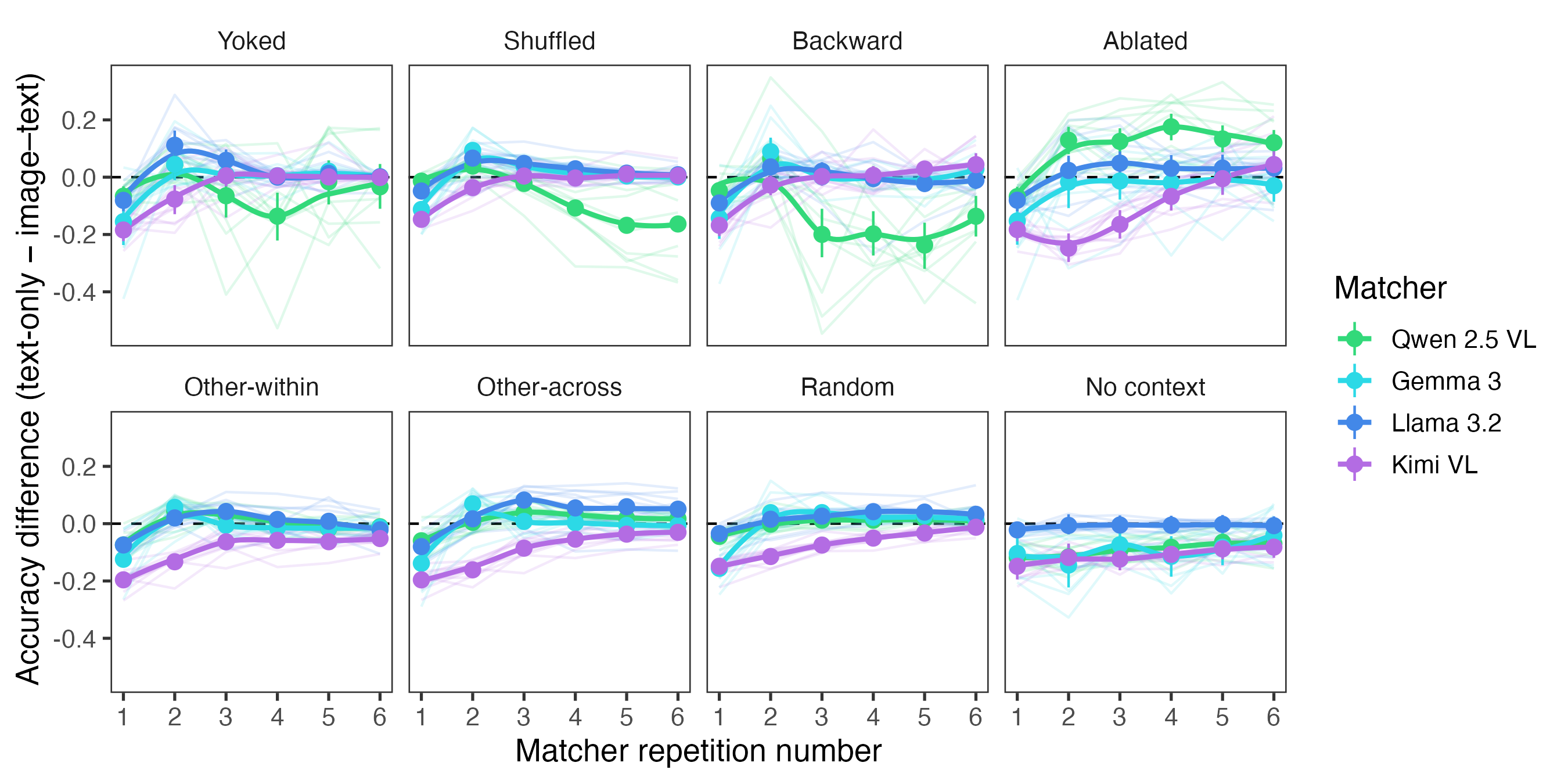}
    \caption{Difference in accuracy between the text-only and image--text paradigms across all conditions and models. Error bars indicate bootstrapped 95\% confidence intervals. Dashed lines indicate no difference.}
    \label{fig:diff}
\end{figure}

Previous studies with VLMs have suggested that they underutilise visual information due to attentional issues or bias from prior linguistic knowledge \cite{rahmanzadehgerviVisionLanguageModels2025, chenWhySpatialReasoning2025, voVisionLanguageModels2025}.
To investigate the extent to which VLMs were using the tangram images to perform pragmatic reasoning, we also ran all models using a prompt that did not include the tangram images.
Model accuracies are shown in Figure~\ref{fig:noimg}, and the difference in accuracy between the text-only and the image--text paradigms is shown in Figure~\ref{fig:diff}.

All models showed near- or at-chance performance in the first repetition (across all conditions) and in the no context condition when no images were shown, suggesting that they were in fact using visual information to resolve pragmatic references.
Models also generally performed worse without images than with images, although the change in performance varied across models---notably, Llama 3.2 performance in fact improved in most cases.
This pattern of results suggests that some models (especially Llama 3.2) were primarily using the text information to select the correct match; this finding corroborates results from Hua et al.~\cite{huaTalkLessInteract2024}, showing that when images are shuffled from trial to trial, VLMs exhibit little to no learning over repetitions.

\end{document}